\documentclass[10pt,twocolumn,letterpaper]{article}

\usepackage{cvpr}
\usepackage{times}
\usepackage{epsfig}
\usepackage{graphicx}
\usepackage{amsmath}
\usepackage{amssymb}
\usepackage{subfig}
\usepackage{booktabs}
\usepackage{enumitem}

\usepackage[pagebackref=true,breaklinks=true,letterpaper=true,colorlinks,bookmarks=false]{hyperref}

\cvprfinalcopy 

\begin{document}

\title{Maintaining Discrimination and Fairness in Class Incremental Learning}

\author{Bowen Zhao$^{\dag,\ddag}$\quad Xi Xiao$^{\dag,\ddag}$\quad Guojun Gan$^*$\quad Bin Zhang$^\ddag$\quad Shutao Xia$^{\dag,\ddag}$\\
$^\dag$Tsinghua University\quad $^\ddag$Peng Cheng Laboratory \quad $^*$University of Connecticut\\
{\tt\small zbw18@mails.tsinghua.edu.cn, \{xiaox,xiast\}@sz.tsinghua.edu.cn}\\ {\tt\small bin.zhang@pcl.ac.cn, guojun.gan@uconn.edu}
} 

\maketitle

\begin{abstract}
Deep neural networks (DNNs) have been applied in class incremental learning, which aims to solve common real-world problems of learning new classes continually. One drawback of standard DNNs is that they are prone to catastrophic forgetting. Knowledge distillation (KD) is a commonly used technique to alleviate this problem. In this paper, we demonstrate it can indeed help the model to output more discriminative results within old classes. However, it cannot alleviate the problem that the model tends to classify objects into new classes, causing the positive effect of KD to be hidden and limited. We observed that an important factor causing catastrophic forgetting is that the weights in the last fully connected (FC) layer are highly biased in class incremental learning. In this paper, we propose a simple and effective solution motivated by the aforementioned observations to address catastrophic forgetting. Firstly, we utilize KD to maintain the discrimination within old classes. Then, to further maintain the fairness between old classes and new classes, we propose Weight Aligning (WA) that corrects the biased weights in the FC layer after normal training process. Unlike previous work, WA does not require any extra parameters or a validation set in advance, as it utilizes the information provided by the biased weights themselves. The proposed method is evaluated on ImageNet-1000, ImageNet-100, and CIFAR-100 under various settings. Experimental results show that the proposed method can effectively alleviate catastrophic forgetting and significantly outperform state-of-the-art methods.
\end{abstract}

\section{Introduction}

\begin{figure}[t]
  \centering
  \includegraphics[width=0.42\textwidth]{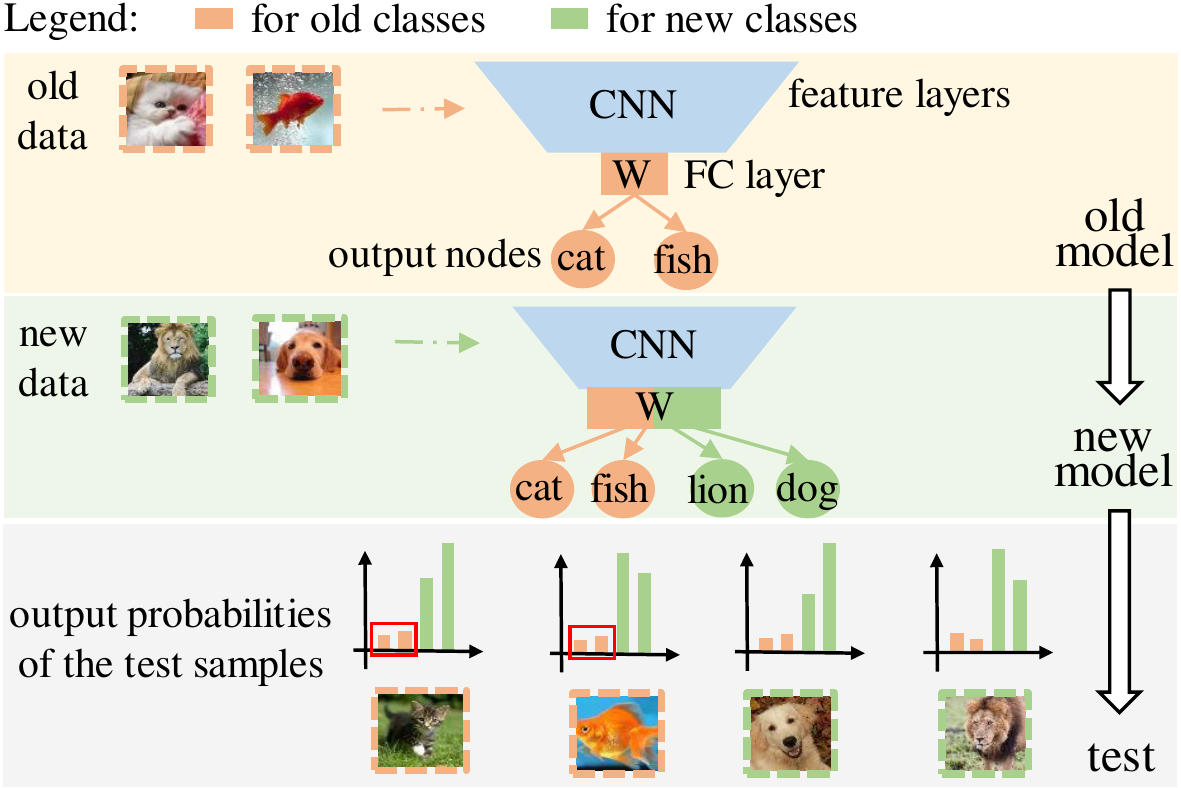}
  \caption{Vanilla  method for class incremental learning.}
  \label{fig:introduction}
\end{figure}

In the past few years, Deep Neural Networks (DNNs) have shown remarkable performance in various applications, even surpassing human performance on some tasks \cite{He2015DeepRL,He2016IdentityMI,Huang2016DenselyCC}. The standard DNNs are typically trained on a prepared dataset, where the number of categories is fixed in advance. However, in many real-world applications, it is often required to learn new classes gradually from streaming data, which is called class incremental learning.

In order to achieve this goal, a common method is to fine tune the old model on new data by setting the number of output nodes to be that of current classes (including old and new classes) as shown in Figure \ref{fig:introduction}. However, this naive method suffers from a serious problem known as catastrophic forgetting \cite{French1999CatastrophicFI,MichaelCatastrophic}. As can be seen from Figure \ref{fig:introduction}, the old data's output probabilities corresponding to the old classes (which are highlighted in red boxes) are relatively low. Thus, the new model trained by the vanilla method generally predicts objects as new classes \cite{rebuffi2017icarl,wu2019large,Zhang2019LabelMN}.

To alleviate catastrophic forgetting, many studies have been done. EWC \cite{kirkpatrick2017overcoming}, SI \cite{zenke2017continual}, and MAS \cite{aljundi2018memory} attempt to solve this problem with a parameter control strategy. Knowledge distillation (KD) \cite{hinton2015distilling} is another strategy, which has also been widely used in this field \cite{castro2018end,li2017learning,zhou2019m2kd}. Besides, some other studies \cite{Nguyen2017VariationalCL,rebuffi2017icarl,shin2017continual,Wu2018IncrementalCL} follow a rehearsal strategy by using a small amount of real or generated old data in the training process. In class incremental learning tasks, the new model is trained without access to the old data, even with the rehearsal strategy, the training set in an incremental step is seriously imbalanced between old classes and new classes. Thus, there are also some studies that deal with catastrophic forgetting from this perspective \cite{hou2019learning,wu2019large}.

\begin{figure}[t]
  \centering
  \includegraphics[width=0.42\textwidth]{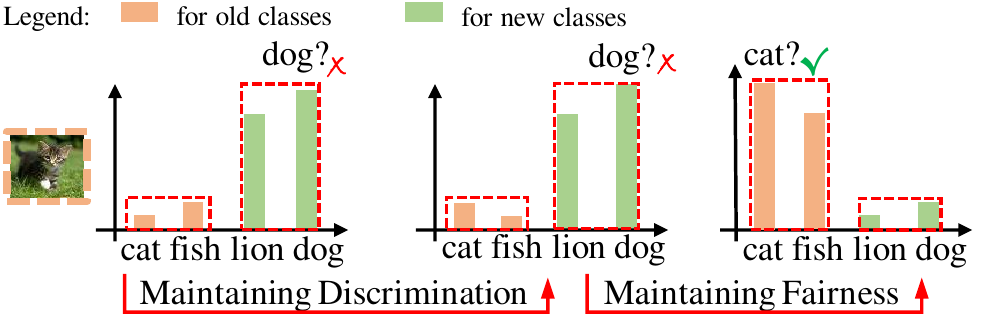}
  \caption{The effect of our solution. KD helps model to maintain discrimination within old classes. WA helps model to maintain fairness between old and new classes.}
  \label{fig:effect}
\end{figure}

In this paper, we first demonstrate that knowledge distillation, the commonly used technique in this field, can indeed help the model to output more discriminative results within old classes. However, the prediction bias towards new classes cannot be alleviated. The trained model still treats old classes unfairly, causing the positive effect of KD to be hidden and limited. Then we show that the weights in the trained model's FC layer are heavily biased, which can cause the model to tend to classify samples into new classes. Based on the above, we present a simple and effective solution to mitigate catastrophic forgetting. The effect of our solution is presented in Figure \ref{fig:effect}. Firstly, we utilize KD to maintain the discrimination within old classes. This helps the model to output more discriminative results within old classes. Then, to further maintain the fairness between old and new classes, we propose Weight Aligning (WA) that corrects the biased weights in the FC layer after the normal training process. This helps the model to treat old classes and new classes fairly, and output correct predictions.

In this paper, our main contributions are the following: 
\begin{itemize}[noitemsep]
  \item[(i)] We investigated the actual role of KD in class incremental learning by experiments, including positive and negative impacts; 
  \item[(ii)] We presented a simple and effective solution to address catastrophic forgetting in class incremental learning that maintains both the discrimination via KD and the fairness via WA; 
  \item[(iii)] Inspired by a prior observation of a non-incremental model, the proposed method WA attempts to align the norms of the weight vectors for new classes to those for old classes. WA makes full use of the information contained in the trained model and correct the biased weights in the FC layer, it does not need to reserve a validation set in advance or require any additional parameters to be tuned, but can handle class incremental learning tasks well; 
  \item[(iv)] Extensive experiments were conducted, the results show that our method achieves better performance than previous methods.  
\end{itemize}

\section{Related Work}\label{sec:related}
Recently, many methods have been proposed to alleviate the well-known problem of catastrophic forgetting \cite{French1999CatastrophicFI,MichaelCatastrophic} suffered by ordinary DNNs . In this section, we briefly discuss these methods.

\textbf{Parameter Control.} The approaches of this strategy such as EWC \cite{kirkpatrick2017overcoming}, SI \cite{zenke2017continual}, and MAS \cite{aljundi2018memory} manage to constrain the important weights of old model when facing new data. These methods expect small changes in the important parameters. They differ in how to estimate the important parameters. EWC estimates the weight importance through the Fisher information matrix; SI uses the path integral over the optimization trajectory; MAS utilizes the gradients of the network output \cite{zhang2019class}. However, the importance of parameters is difficult to measure accurately in a series of tasks \cite{hou2019learning}. These methods tend to perform poorly in class incremental learning \cite{hsu2018re,DBLP:journals/corr/abs-1904-07734}.

\textbf{Knowledge Distillation.} Knowledge distillation \cite{hinton2015distilling} is a widely used method, which transfers key knowledge from a teacher model to a student model. LwF \cite{li2017learning} utilizes knowledge distillation to learn multiple tasks. A modified cross-entropy loss is used to preserve the capabilities of old model. Then, it was applied to multi-class classification, called LwF.MC \cite{rebuffi2017icarl}. $M^2$KD \cite{zhou2019m2kd} introduces a multi-model and multi-level knowledge distillation strategy, which utilizes all previous model snapshots instead of distilling knowledge only from the last model.

\textbf{Rehearsal.} The rehearsal strategy alleviates catastrophic forgetting by using some old data to make up training data. The simplest approach is to store few old data and replay them in a new incremental step. This straightforward approach has been demonstrated to be effective in many scenarios \cite{hsu2018re,DBLP:journals/corr/abs-1904-07734}. Other methods construct a generative model, e.g., GANs \cite{goodfellow2014generative}, to generate samples for rehearsal instead of storing old data directly \cite{farquhar2019unifying,shin2017continual,Wu2018IncrementalCL}. However, in these methods, an additional generative model needs to be trained simultaneously. Therefore, they rely heavily on the quality of the generated model.

\textbf{Class Imbalance.} For class incremental learning, data of old classes is generally not available when new classes appear. Even with the rehearsal strategy, the class imbalance problem is still very serious, which is an important factor in catastrophic forgetting \cite{hou2019learning,wu2019large}. Though class imbalance is an old topic and has attracted a lot of attention \cite{Buda2018ASS,Huang2016LearningDR,Khan2015CostSensitiveLO}, multi-class imbalance learning is still an open problem \cite{Zhang2019MultiImbalanceAO}. In order to address it in class incremental learning, BiC \cite{wu2019large} adds a bias correction layer to correct the model's outputs. This method needs to keep a validation set to train the additional bias correction layer. In \cite{hou2019learning}, cosine normalization, less-forget constraint, and inter-class separation are incorporated to mitigate the impact of class imbalance. This method combines three specific loss terms and other skills (e.g., class balance fine tune) to improve performance. IL2M \cite{Belouadah_2019_ICCV} rectifies scores of old classes by leveraging contents from a dual memory.

These strategies can be applied in combination. For example, both the distillation strategy and the rehearsal strategy are used in iCaRL \cite{rebuffi2017icarl}, which also utilizes a nearest-exemplars-mean (NEM) classifier. EEIL \cite{castro2018end} also exploits these two strategies and utilizes a balanced fine tuning to alleviate class imbalance. In this paper, the proposed method is also based on these perspectives. A detailed analysis of distillation strategy is presented, including its positive and negative effects. More importantly, we deal with class imbalance in a simple and effective manner. Without any additional model parameters, hyperparameters or a reserved validation set, our method achieves better performance than previous methods.

\section{Motivation}\label{sec:motivation}

\subsection{Baseline}
In this subsection, we summarize a baseline method in class incremental learning, which utilizes both the rehearsal strategy and the distillation strategy. 

Let us first formulate class incremental learning. Assume there are $B$ batches of train data $\{D^1, \cdots, D^B\}$, with $D^b = \{(\mathbf{x}^b_1, y^b_1), \cdots, (\mathbf{x}^b_{n_b}, y^b_{n_b})\}$ for the $b^{th}$ incremental step, where $\mathbf{x}^b_i$ and $y^b_i$ represent the input data and the target respectively, $n_b$ is the number of samples in the set $D^b$. In the $b^{th}$ step of class incremental learning, the goal is to learn knowledge from new data $D^b$, while retain the previous experiences learned from old data $\{D^1, \cdots, D^{b-1}\}$. For each step, the trained model is evaluated on all seen classes.

For the $b^{th}$ incremental step, the baseline method initializes the model with the parameters learned in the previous step and adds new output nodes (weights in the FC layer are initialized randomly). Then, it attempts to learn new classes and meanwhile preserve the original capabilities with the new data $D^b$ and a few rehearsal data $D^b_{old}$. It is assumed that the new data $D^b$ comes from $C^b$ new classes, and the rehearsal data $D^b_{old}$ comes from $C^b_{old}$ old classes, where $C^b_{old} = \sum_{k=1}^{b-1} C^k$. The baseline method combines the cross-entropy loss $\mathcal{L}_{CE}$ with the knowledge distillation loss $\mathcal{L}_{KD}$. The combined loss containing two terms is given as:
\begin{equation}
  \mathcal{L} (\mathbf{x}, y) = (1-\lambda) \mathcal{L}_{CE} (\mathbf{x}, y) + \lambda \mathcal{L}_{KD} (\mathbf{x}),
  \label{eq:combine_loss}
\end{equation}
where $\lambda$ is a hyper-parameter governing the balance between the two losses. We set the hyper-parameter $\lambda$ to $\frac{C^b_{old}}{C^b+C^b_{old}}$, according to the recommendation in \cite{wu2019large}. The cross-entropy loss is given by:
\begin{equation}
  \mathcal{L}_{CE} (\mathbf{x}, y) = 
  \sum_{c=1}^{C^b+C^b_{old}}
  - \delta_{c=y} \log\big( p_c(\mathbf{x}) \big),
  \label{celoss}
\end{equation}
where $\delta_{c=y}$ is the indicator function and $p_c(\mathbf{x})$ is the output probability for the $c^{th}$ class. And the distillation loss is given by:
\begin{equation}
  \mathcal{L}_{KD} (\mathbf{x}) = 
  \sum_{c=1}^{C^b_{old}}
  - \hat{q}_c(\mathbf{x}) \log\big( q_c(\mathbf{x}) \big),
  \label{distillationloss}
\end{equation}
where 
$
\hat{q}_c(\mathbf{x}) = \frac{ e^{\hat{o}_c(\mathbf{x})/T}}{ \sum_{j=1}^{C^b_{old}} e^{\hat{o}_j(\mathbf{x})/T} },
q_c(\mathbf{x}) = \frac{ e^{o_c(\mathbf{x})/T}}{ \sum_{j=1}^{C^b_{old}} e^{o_j(\mathbf{x})/T} };
$
$T$ is the temperature scalar; $\hat{o}_c(\mathbf{x})$ is an element of $\hat{\mathbf{o}}(\mathbf{x})$, $\hat{\mathbf{o}}(\mathbf{x}) = \big(\hat{o}_1(\mathbf{x}), \cdots, \hat{o}_{C^b_{old}}(\mathbf{x})\big)^T$, which represents the output logits of the old model obtained in the previous incremental step; $o_c(\mathbf{x})$ is an element of $\mathbf{o}(\mathbf{x})$,
$\mathbf{o}(\mathbf{x}) = \big(o_1(\mathbf{x}), \cdots, o_{C^b_{old}}(\mathbf{x}), o_{C^b_{old}+1}(\mathbf{x}), \cdots, o_{C^b_{old}+C^b}(\mathbf{x})\big)^T$, which stands for the output logits of the current model. Note the sample $(\mathbf{x}, y)$ is from both the new data and the rehearsal data. Then, parameters of both the feature extraction layers and the FC layer are updated with the combined loss defined in Eq.(\ref{eq:combine_loss}) during training. 

\begin{table}[t]
\centering
\caption{Error analysis on two parts of the test set. $e(o)$, $e(n)$ represent the number of old samples and new samples that are wrongly predicted, respectively. Specifically, error analysis for old samples is given in detail: $e(o, n)$, $e(o, n)$ stand for the number of old samples that are misclassified as new classes or other old classes, respectively.}

\begin{tabular}{c|c|c|c|c}
  \toprule
        & $e(n)$ & $e(o)$ & $e(o,n)$ & $e(o,o)$ \\
  \midrule
  \midrule
  CE    & 314   & 5,360 & 4,027 & 1,333 \\
  CE + KD & 383   & 5,326 & 4,314 & 1,012 \\
  \bottomrule
  \end{tabular}%
\label{tab:fivesteps_results}%
\end{table}

\begin{figure*}[t]
\centering
\includegraphics[width=0.9\textwidth]{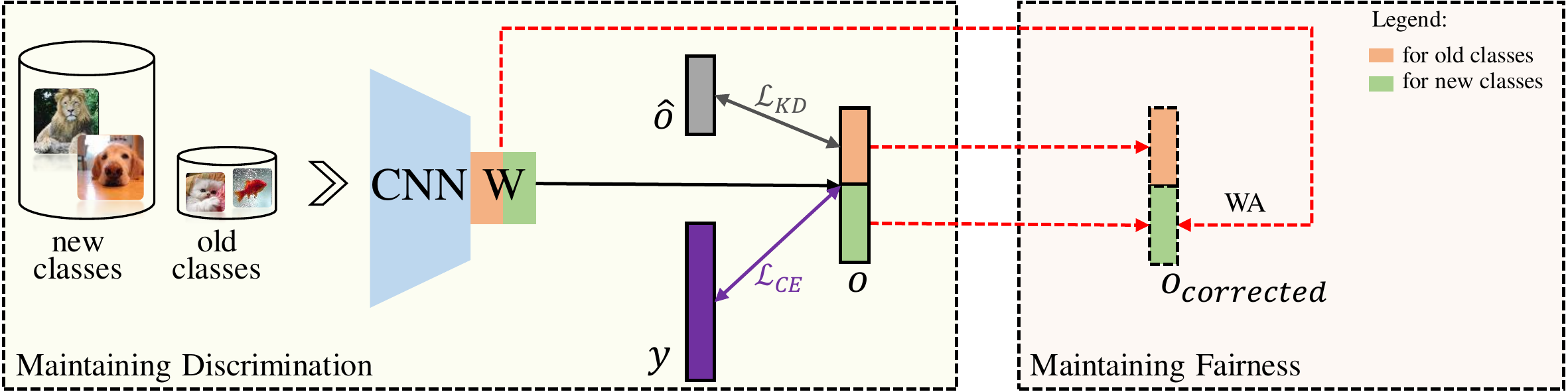}
\caption{Overview of our solution for class incremental learning. In the first phase, we train the model with the cross-entropy loss ($\mathcal{L}_{CE}$) and the distillation loss ($\mathcal{L}_{KD}$). In the second phase, we correct the biased weights in the trained model via Weight Aligning (WA). $\mathbf{o}$ and $\mathbf{\hat{o}}$ represent the output logits of the current model and the old model respectively, $y$ stands for the true label, $\mathbf{o}_{corrected}$ represents the corrected output logits by using WA.}
\label{fig:method}
\end{figure*}  

\subsection{Effect of Knowledge Distillation}\label{sec:e_kd}
The baseline method is widely used in class incremental learning. However, there is a lack of explicit analysis of the role of knowledge distillation. To do this, we carry out experiments on the CIFAR-100 \cite{Krizhevsky09} with 5 incremental steps ($B=5$) and 20 classes per step ($C^b=20$, $b=1,\cdots,5$). 

We perform class incremental learning with two methods: (a) using the cross-entropy loss; (b) using both the cross-entropy loss and the distillation loss. After 5 incremental steps, we evaluate the two models trained by method (a) and (b). The test set is comprised of two parts, one containing 80 old classes and another 20 new classes. Error analysis on two parts of the test set is reported in Table \ref{tab:fivesteps_results}. There are 2,000 test samples in the new part, and 8,000 samples in the old part. As can be seen, both methods have very poor performance in term of old classes, which shows that they have lost the ability to recognize old data.

We further analyze the type of misclassification of old data. As shown in Table \ref{tab:fivesteps_results}, the combined loss reduces the number of old samples that are misclassified to other old classes: 1,012 (CE + KD) vs 1,333 (CE). This is consistent with the original intention of knowledge distillation, that is, to keep the knowledge of old model. 
However, the prediction bias towards new classes is not alleviated: there are more old samples that are misclassified to new classes: 4,314 (CE + KD) vs 4,027 (CE). Why dose the model trained with the distillation loss become more serious towards new classes? After revisiting the distillation loss, we find the cost of misclassifying old samples to new classes is smaller than that to other old classes. If old samples are misclassified to new classes, the distillation loss still can be low, as \{$q_c(\mathbf{x}), \ c=1,\cdots,C^b_{old}$\} are only calculated between the outputs corresponding to old classes. While, if they are misclassified to other old classes, the distillation loss will be high, as the output probability distribution is definitely not coincide with the target distribution. As a result, the model is more inclined to misclassify old samples into new classes.

Based on the above analysis, we argue that the positive effect of distillation loss is maintaining the discrimination within old classes, so that it is successful in making fewer misclassifications within old classes. However, the model still has a prediction bias towards new classes. The positive effect of knowledge distillation here is limited. Besides, if there are more than two incremental steps, i.e., $B>2$, the `ill' model will become a teacher model in the next incremental step, then the deviation will accumulate, so that the positive effect will be further limited.

\section{Methodology}\label{sec:method}
Our method consists of two phases, as shown in Figure \ref{fig:method}. The first phase is \textbf{Maintaining Discrimination}. In this phase, we train a new model on the new data and the rehearsal data with the combined loss. We expect to transfer knowledge from the old model to the new model and maintain discrimination within old classes with the help of knowledge distillation. 

As knowledge distillation loss still cannot help the model to treat old classes and new classes fairly as shown in subsection \ref{sec:e_kd}, we design the second phase, called \textbf{Maintaining Fairness}. In this phase, we propose a method named Weight Aligning (WA) to correct the model trained in the first phase. The corrected model treats old classes and new classes fairly, and can significantly improve the overall performance.

\subsection{Biased Weights in the FC Layer}\label{sec:biased_weights}

\begin{figure*}[t]
\centering
\subfloat[$C^1=20, C^1_{old}=0$]{\includegraphics[width=0.195\textwidth]{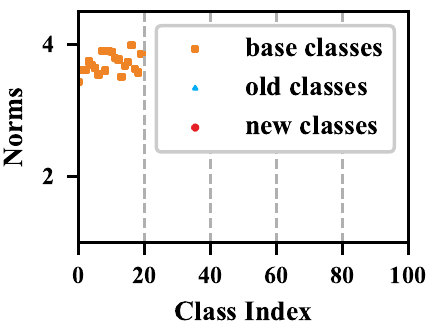}}
\subfloat[$C^2=20, C^2_{old}=20$]{\includegraphics[width=0.195\textwidth]{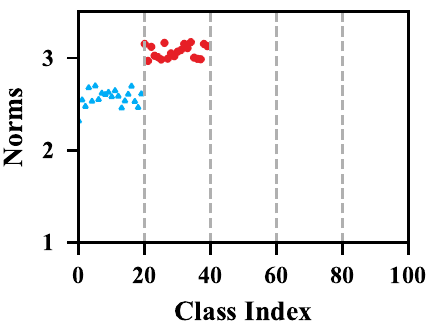}}
\subfloat[$C^3=20, C^3_{old}=40$]{\includegraphics[width=0.195\textwidth]{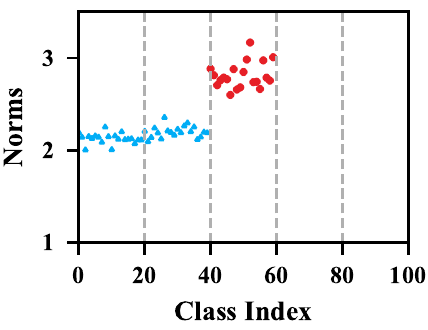}}
\subfloat[$C^4=20, C^4_{old}=60$]{\includegraphics[width=0.195\textwidth]{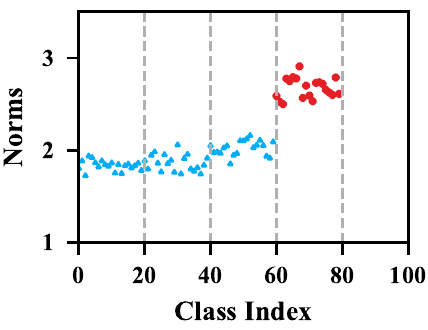}}
\subfloat[$C^5=20, C^5_{old}=80$]{\includegraphics[width=0.195\textwidth]{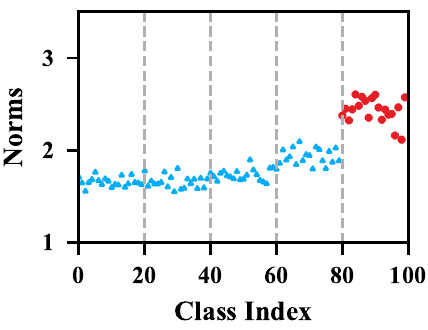}}
\caption{Norms of the weight vectors $\{\mathbf{w}_c\}$. (a) Results of the $1^{st}$ step (20 base classes), which does not correspond to class incremental learning; (b), (c), (d) and (e) are the results of the $2^{nd}$, $3^{rd}$, $4^{th}$, $5^{th}$ incremental step respectively, which show the norms of the weight vectors of new classes are much larger than those of old classes. (Best viewed in color)}
\label{fig:norms}
\end{figure*}

As shown in subsection \ref{sec:e_kd}, the model trained via the baseline method still tends to predict test samples as new classes. To study this problem conveniently, we express the FC layer of model in the $b^{th}$ incremental step in the following form:
\begin{equation}
  \mathbf{o}(\mathbf{x}) = \mathbf{W}^T \phi(\mathbf{x}),
\end{equation}
where the $(C^b_{old}+C^b)$-dimensional vector $\mathbf{o}(\mathbf{x})$ represents output logits of the current model; $\phi(\cdot)$ is a feature extraction function (can be a CNN-based model usually), which outputs $d$-dimensional feature vectors; $\mathbf{W} \in \mathbb{R}^{d\times (C^b_{old}+C^b)}$ stands for the weights, which can be expressed as $\mathbf{W} = \{\mathbf{w}_c, 1 \leq c \leq C^b_{old}+C^b\}$, where $\mathbf{w}_c$ is a $d$-dimensional weight vector for the $c^{th}$ class. Note, for the convenience of analysis, we always set the bias term in the FC layer to zero without special instructions, which will be discussed in the ablation study.

We carry out experiments on CIFAR-100 with 5 incremental steps and 20 classes per step. After each step, we calculate the norms of the weight vectors $\{\mathbf{w}_c\}$ and plot them in Figure \ref{fig:norms}. As shown in Figure \ref{fig:norms} (b), (c), (d) and (e), the norms of the weight vectors for new classes are much larger than those for old classes. This phenomenon is mainly caused by class imbalance \cite{Guo2017OneshotFR,Liu2017SphereFaceDH}. Due to the output logits for the $c^{th}$ class is calculated as 
\begin{equation}
  o_c(\mathbf{x})=\mathbf{w}_c^T \phi(\mathbf{x}),
  \label{eq:o_c}
\end{equation}
if the norms of weight vectors for new classes are larger, the output logits for new classes may tend to be larger in general. As a result, the trained model may tend to predict an input image as belonging to a new class. 

However, as shown in Figure \ref{fig:norms} (a), in the first phase, the norms of the weight vectors are roughly equal, as this phase does not related to class incremental learning actually. We treat this as a priori knowledge. The phenomenon in class incremental learning does not match this prior knowledge, which inspires us to correct the biased weights.

\subsection{Weight Aligning}\label{sec:WA}
Based on the above, we present a simple and effective approach, called Weight Aligning (WA), to correct the biased weights in the FC layer. In WA, the norms of the weight vectors of new classes are aligned to those of old classes.

Firstly, we rewrite the weights in the FC layer in the following form
$$
\mathbf{W} = (\mathbf{W}_{old}, \mathbf{W}_{new}),
$$
where 
\begin{equation*}
  \begin{aligned}
     &\mathbf{W}_{old} = ( \mathbf{w}_1, \mathbf{w}_2, \cdots, \mathbf{w}_{C^b_{old}} ) \in \mathbb{R}^{d\times C^b_{old}} ,\\
     &\mathbf{W}_{new}= ( \mathbf{w}_{C^b_{old}+1},\cdots,\mathbf{w}_{C^b_{old}+C^b} ) \in \mathbb{R}^{d\times C^b} .
  \end{aligned}
\end{equation*}
Then, we denote, respectively, the norms of the weight vectors of old classes and new classes  as follows
\begin{equation*}
  \begin{aligned}
  &\textit{\textbf{Norm}}_{old} = (||\mathbf{w}_1||, \cdots, ||\mathbf{w}_{C^b_{old}}||),\\
  &\textit{\textbf{Norm}}_{new} = (||\mathbf{w}_{C^b_{old}+1}||, \cdots, ||\mathbf{w}_{C^b_{old}+C^b}||) .
  \end{aligned}
\end{equation*}

Based on the above norms, we normalize the weights for new classes by
\begin{equation}
\widehat{\mathbf{W}}_{new} = \gamma \cdot \mathbf{W}_{new} ,
\label{eq:normalize}
\end{equation}
where 
\begin{equation}
  \gamma = \frac{Mean(\textit{\textbf{Norm}}_{old})}{Mean(\textit{\textbf{Norm}}_{new})},
  \label{eq:gamma}
\end{equation}
$Mean(\cdot)$ returns the mean value of elements in the vector.
In this way, the average norm of the weight vectors for new classes becomes the same as that for old classes. Note that we only make the average norms become equal, in other words, within new classes (or old classes), the relative magnitude of the norms of the weight vectors does not change. Such a design is mainly used to ensure the data within new classes (or old classes) can be separated well.

The original output logits of the model trained in the first phase of our method can be expressed as 
\begin{equation}
  \begin{aligned}
    \mathbf{o}(\mathbf{x}) &= 
    \big(
      \mathbf{o}_{old}(\mathbf{x}),\
      \mathbf{o}_{new}(\mathbf{x})
    \big)^T\\
    &=
    \big(
      \mathbf{W}_{old}^T\ \phi(\mathbf{x}),\
      \mathbf{W}_{new}^T\ \phi(\mathbf{x})
    \big)^T.
  \end{aligned}
\end{equation}
After applying WA to the weights, the corrected output logits are given by:
\begin{equation}
  \begin{aligned}
  \mathbf{o}_{corrected}(\mathbf{x}) &= 
  \big(
      \mathbf{W}_{old}^T \ \phi(\mathbf{x}),\
      \widehat{\mathbf{W}}_{new}^T \ \phi(\mathbf{x})
  \big)^T
  \\
  &=
  \big(
      \mathbf{W}_{old}^T \ \phi(\mathbf{x}),\
      \gamma \cdot \mathbf{W}_{new}^T \ \phi(\mathbf{x})
  \big)^T
  \\
  &=
  \big(
      \mathbf{o}_{old}(\mathbf{x}),\
      \gamma \cdot \mathbf{o}_{new}(\mathbf{x})
  \big)^T.
  \end{aligned}
  \label{eq:rescale}
\end{equation}
As shown in Eq.(\ref{eq:rescale}) and Eq.(\ref{eq:gamma}), the final effect of aligning the weights is to rescale the output logits of new classes by a coefficient. The latter experiments demonstrate that our method can effectively alleviate the prediction bias.

\begin{table*}[t]
  \centering
  \caption{Class incremental learning performance (top-1 accuracy \%) on CIFAR-100 with 5 incremental steps and 20 classes per step. The gains on the basis of Variation1 are also reported in parentheses. The upper bound performance is obtained with all training data for all classes. The average results over all the incremental steps except the first step are also reported here (as the first step does not related to class incremental learning actually). The best results are in bold.}
\resizebox{\textwidth}{19mm}{
\begin{tabular}{c|ccccccccccc}
  \toprule
  \#classes & \multicolumn{1}{c|}{20 } & \multicolumn{2}{c|}{40 } & \multicolumn{2}{c|}{60 } & \multicolumn{2}{c|}{80 } & \multicolumn{2}{c|}{100 } & \multicolumn{2}{c}{Average} \\
  \midrule
  \midrule
  Variation1 (CE) & \multicolumn{1}{c|}{83.5 } & 70.7  & \multicolumn{1}{c|}{} & 58.2  & \multicolumn{1}{c|}{} & 49.2  & \multicolumn{1}{c|}{} & 43.3  & \multicolumn{1}{c|}{} & 55.3  &  \\
  Variation2 (CE + WA) & \multicolumn{1}{c|}{83.5 } & 74.3  & \multicolumn{1}{c|}{(+3.6)} & 64.0  & \multicolumn{1}{c|}{(+5.8)} & 56.9  & \multicolumn{1}{c|}{(+7.7)} & 50.8  & \multicolumn{1}{c|}{(+7.5)} & 61.5  & (+6.2) \\
  Variation3 (CE + KD) & \multicolumn{1}{c|}{83.5 } & 72.8  & \multicolumn{1}{c|}{(+2.1)} & 60.1  & \multicolumn{1}{c|}{(+1.9)} & 49.9  & \multicolumn{1}{c|}{(+0.7)} & 42.9  & \multicolumn{1}{c|}{(-0.4)} & 56.4  & (+1.1) \\
  Variation4 (CE + KD + WNL) & \multicolumn{1}{c|}{83.1 } & 72.3  & \multicolumn{1}{c|}{(+1.6)} & 61.6  & \multicolumn{1}{c|}{(+3.4)} & 53.1  & \multicolumn{1}{c|}{(+3.9)} & 46.0  & \multicolumn{1}{c|}{(+2.7)} & 58.2  & (+2.9) \\
  Ours (CE + KD + WA) & \multicolumn{1}{c|}{83.5 } & \textbf{75.5} & \multicolumn{1}{c|}{(+4.8)} & \textbf{68.7} & \multicolumn{1}{c|}{(+10.5)} & \textbf{63.1} & \multicolumn{1}{c|}{(+13.9)} & \textbf{59.2} & \multicolumn{1}{c|}{(+15.9)} & \textbf{66.6} & (+11.3) \\
  \midrule
  Upper Bound  & \multicolumn{11}{c}{70.1 } \\
  \bottomrule
  \end{tabular}%
}
\label{tab:ablation}%
\end{table*}%

\subsection{Restriction to the Weights}\label{sec:r_to_w}
In fact, the magnitude relationship between the norms of weight vectors for new classes and those for old classes may not always reflect the magnitude relationship between the output logits for old classes and those for new classes. Suppose that the feature extraction function provides the feature vectors, whose elements are all non-negative. This assumption is reasonable, because in usual model architectures, the learned features are activated by the `ReLU' function $\big( \text{ReLU}(x)=\max(0,x) \big)$, which returns non-negative values. As the weight vectors $\{\mathbf{w}_c\}$ usually contain both positive and negative elements, the negative elements with large absolute values contribute to a large norm of weight vectors. However, they are not in favor of large output logits. 
Thus, in order to make the norm of the weight vector $\mathbf{w}_c$ more consistent with its corresponding output logits, we restrict the elements of the weight vector $\mathbf{w}_c$ to be positive. To achieve this, weight clipping \cite{Arjovsky2017WassersteinG} can be performed after each optimization step in training. The impact of restricting the weights in the FC layer to be positive will be analyzed in the ablation study.

\section{Experiments}\label{sec:experiment}

\subsection{Experimental Settings}
We evaluate the methods on ImageNet ILSVRC 2012 \cite{Russakovsky2015} and CIFAR-100 \cite{Krizhevsky09}, which are widely used in the study of class incremental learning \cite{castro2018end,rebuffi2017icarl,wu2019large}. ImageNet ILSVRC 2012 is a large-scale dataset with 1,000 classes that includes about 1.2 million images for training and 50,000 images for validation. CIFAR-100 consists $32\times 32$ pixel color images with 100 classes. It contains 50,000 images for training with 500 images per class, and 10,000 images for evaluating with 100 images per class.

Our method are implemented with Pytorch \cite{paszke2017automatic}. The code will be made publicly available. For ImageNet, we adopt a 18-layer ResNet \cite{He2015DeepRL,He2016IdentityMI}. We use SGD to train our model and set the batch size to 256. The learning rate starts from 0.1 and reduces to 1/10 of the previous learning rate after 30, 60, 80 and 90 epochs (100 epochs in total). For CIFAR-100, we use a 32-layer ResNet. We also train the model with SGD and set the batch size to 32. The learning rate starts from 0.1 and reduces to 1/10 of the previous learning rate after 100, 150 and 200 epochs (250 epochs in total). We set the temperature scalar $T$ to 2. For data augmentation, random cropping, horizontal flip and normalization are employed to augment training images. 

\subsection{Effect of Weight Aligning}
To analyze the effect of weight aligning, we perform experiments on CIFAR-100 with 5 incremental steps and 20 classes per step. We first compare our method with three variations in the following: \textbf{Variation1}, training with the cross-entropy loss; \textbf{Variation2}, training with the cross-entropy loss, and correcting the model via WA; \textbf{Variation3}, training with the combined loss; \textbf{Ours}, training with the combined loss and correcting the model via WA.

Table \ref{tab:ablation} summarizes the results of these experiments. Variation1 is the worst one, as it only uses the cross-entropy loss. Variation3 adds the distillation loss on the basis of Variation1 to mitigate catastrophic forgetting. However, Variation3 is only a little better than Variation1. Variation2 uses WA to correct the model based on Variation1, and significantly improves performance (the gain in term of the overall performance at the end of class incremental learning is 7.5\%). From the results of `Ours', WA also gets significant improvements (more than 16\% at the end of class incremental learning over Variation3). These results demonstrate that WA is quite effective for class incremental learning. 

It is worth noting that the gain brought by the combination of KD and WA is greater than the sum of the gains from each component used separately, e.g., for the average results, the gain of the combination (Ours) is 11.3\%, and the gains of WA (Variation2) and KD (Variation3) used separately are 6.2\% and 1.1\% respectively. As shown in subsection \ref{sec:e_kd}, the positive effect of KD is limited when used alone. KD helps the model to output more discriminative results within old classes, however, these outputs are overwhelmed by the superior outputs of new classes. For example, as shown in Figure \ref{fig:effect}, with the help of KD, the output probability for `cat' becomes higher than that for `fish', but still lower than that for new class `lion' or `dog'. In such a scenario, the positive effect of KD is hidden. As our method maintains not only the discrimination within old classes but also the fairness between old classes and new classes, it strengthens the positive effect of KD. On the other hand, the corrected outputs via WA are more accurate with the help of KD. Therefore, our method creates the ``one plus one greater than two" effect and achieves significant improvements.

\begin{figure}[t]
  \centering
  \subfloat[Variation1 (CE)]{\includegraphics[width=0.2\textwidth]{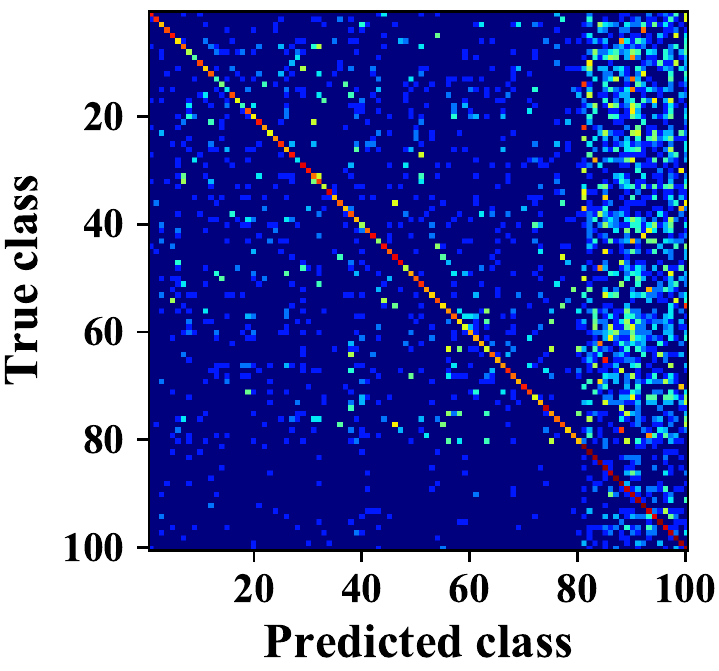}}\hspace{5mm}
  \subfloat[Variation2 (CE + WA)]{\includegraphics[width=0.2\textwidth]{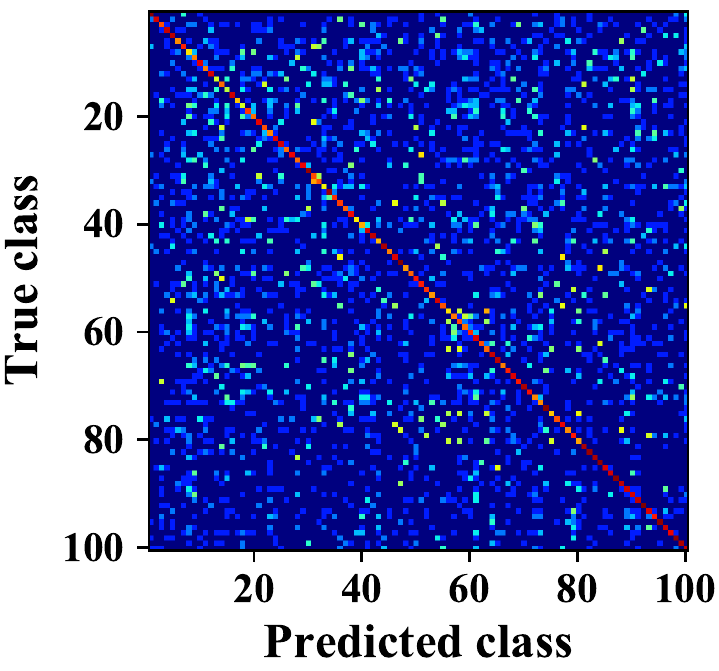}}\hspace{5mm}\\ \vspace{-3mm}
  \subfloat[Variation3 (CE + KD)]{\includegraphics[width=0.2\textwidth]{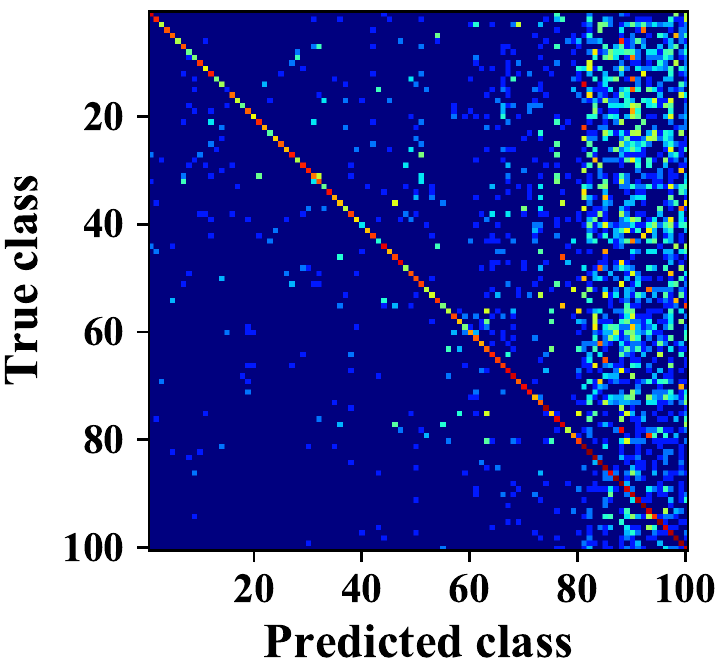}}\hspace{5mm}
  \subfloat[Ours (CE + KD + WA)]{\includegraphics[width=0.2\textwidth]{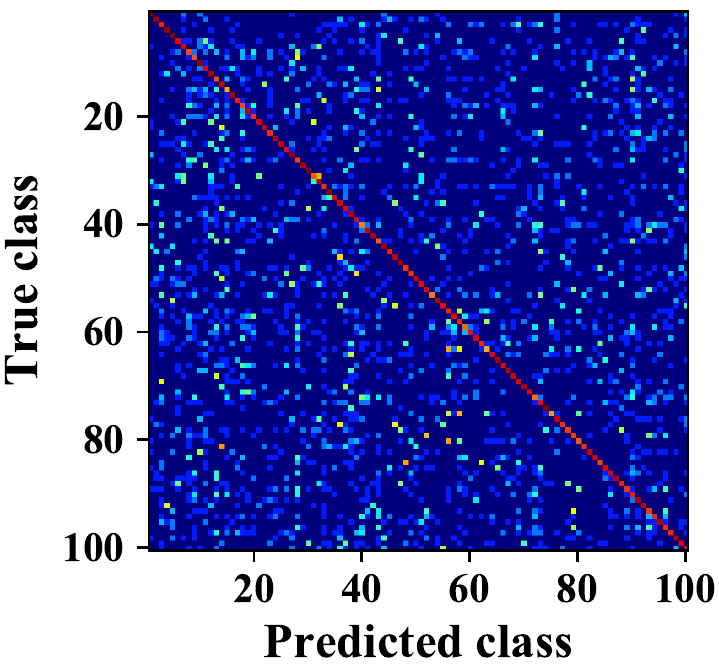}}
  \caption{Confusion matrices (with entries transformed by $\log(1+x)$ for better visibility) of different approaches.}
  \label{fig:confusion}
\end{figure}

The confusion matrices of different methods are presented in Figure \ref{fig:confusion}. From Figure \ref{fig:confusion} (a) and (c), we see that KD leads to fewer misclassifications between old classes, however, both Variation1 and Variation3 tend to predict objects as new classes. With the help of WA, Variation2 and our method make the model treat new classes and old classes fairly as shown in Figure \ref{fig:confusion} (b) and (d). And our method achieves better performance with the help of KD. These results intuitively show that the proposed method can effectively maintain discrimination and fairness in the model predictions.

The proposed method weight aligning is a post-processing technique. It is interesting to see the effect of adding a normalization layer on the weights (in the FC layer) directly, like the operation in Modified Softmax Loss \cite{Liu2017SphereFaceDH} and NormFace \cite{Wang2017NormFaceLH}, so that the weights of all classes can have a unit norm. We implement this method as \textbf{Variation4}: training with the combined loss and a weight normalization layer (WNL). The results are also provided in Table \ref{tab:ablation}. Compared with Variation1 and Variation2, this method does not bring about a significant improvement. Actually, the FC layer plays an important role in the visual representation transfer \cite{Zhang2017InDO}. If the weights in the FC layer are strictly limited during the training process, in order to adapt to new data, the bias in the feature extraction layers will become more serious. However, the bias in the feature extraction layers is harder to correct than that in the weights of FC layer, as the parameters of feature extraction layers are shared by all classes and the weights of FC layer are not shared between classes. Therefore, it is better to take a post-processing approach, such as WA. In addition, we have tested the method that normalizing the weights of all classes to have a unit norm after the usual training process. While this approach is inferior to WA. As mentioned in subsection \ref{sec:WA}, within the new classes (or the old classes), the relative magnitude of the norms of the weight vectors does not change in WA, such a design can maintain the differences and ensure that the classes can be separated well.

\subsection{Comparison to Other Methods}
We compare our method with several competitive or representative methods, including LwF.MC \cite{li2017learning,rebuffi2017icarl}, iCaRL \cite{rebuffi2017icarl}, EEIL \cite{castro2018end}, BiC \cite{wu2019large}, IL2M \cite{Belouadah_2019_ICCV}, RPS \cite{Rajasegaran2019RandomPS}. Experiments are performed on ImageNet100, ImageNet1000 and CIFAR100.

\begin{table}[t]
\centering
\caption{Class incremental learning performance (top-5 accuracy \%) on ImageNet (1,000 classes and 100 classes) with 10 incremental steps. The performance at the last incremental step and the average results over all the incremental steps except the first step are reported here. The results of the compared methods are reported in the original papers. The best results are in bold.}
\begin{tabular}{c|cc|cc}
  \toprule
  \#classes & \multicolumn{2}{c|}{1000 } & \multicolumn{2}{c}{100 } \\
  \midrule
        & \multicolumn{1}{c|}{Last} & Average   & \multicolumn{1}{c|}{Last} & Average \\
  \midrule
  \midrule
  LwF.MC \cite{li2017learning,rebuffi2017icarl} & \multicolumn{1}{c|}{24.3 } & 42.5  & \multicolumn{1}{c|}{36.6 } & 60.7  \\
  iCaRL \cite{rebuffi2017icarl} & \multicolumn{1}{c|}{44.0 } & 60.8  & \multicolumn{1}{c|}{63.8 } & 81.8  \\
  EEIL \cite{castro2018end} & \multicolumn{1}{c|}{52.3 } & 69.4  & \multicolumn{1}{c|}{80.2 } & 89.2  \\
  BiC \cite{wu2019large} & \multicolumn{1}{c|}{73.2 } & 82.9  & \multicolumn{1}{c|}{\textbf{84.4}} & 89.8  \\
  IL2M \cite{Belouadah_2019_ICCV} & \multicolumn{1}{c|}{--} & 78.3  & \multicolumn{1}{c|}{--} & -- \\
  RPS \cite{Rajasegaran2019RandomPS} & \multicolumn{1}{c|}{--} & --    & \multicolumn{1}{c|}{74.0 } & 86.6  \\
  Ours  & \multicolumn{1}{c|}{\textbf{81.1}} & \textbf{85.7} & \multicolumn{1}{c|}{84.1 } & \textbf{90.2} \\
  \midrule
  Upper Bound & \multicolumn{2}{c|}{89.1 } & \multicolumn{2}{c}{95.1 } \\
  \bottomrule
  \end{tabular}%
\label{tab:imagenet}%
\end{table}%

\noindent\textbf{Evaluation on ImageNet.}
We conduct two experiments on this dataset. In the first one, 100 classes (ImageNet-100) are selected randomly and split into 10 incremental batches with 10 classes per batch; In the second one, we split the 1000 classes (ImageNet-1000) into 10 incremental batches with 100 classes per batch. For the sake of fairness, we use the same set of classes in ImageNet-100 and ImageNet-1000 as the previous work \cite{wu2019large}. We store 2,000 images for old classes in ImageNet-100 experiments. And in ImageNet-1000 experiments, we store 20,000 images for old classes as the same as the previous work. We select rehearsal exemplars based on herding selection \cite{Welling2009HerdingDW} which is also the same as the previous work. More classes have been seen, fewer images can be retained per class. As a result, the problem of class imbalance becomes more serious. 

The class incremental learning results (top-5 accuracy \%) on ImageNet-100 and Imagenet-1000 are shown in Table \ref{tab:imagenet}. We report the performance at the last incremental step and the average results over all the incremental steps except the first step here (as the first step does not related to class incremental learning actually). We also provide the detailed results of all incremental steps and the top-1 results in the supplementary material. As can be seen from these tables, the proposed method outperforms the compared methods by a large margin, especially on the large scale dataset ImageNet-1000. The overall performance at the end of class incremental learning is improved by more than 28\% compared to EEIL on ImageNet-1000. In contrast to the state-of-the-art method BiC, the proposed method also achieves better results (surpasses it by 7.9\% at the end of class incremental learning on ImageNet-1000). Though Eq.(\ref{eq:rescale}) is similar in form to the linear model in BiC, the proposed method does not need to reserve a validation set which is used in BiC to learn additional parameters. All of the rehearsal data can be utilized to learn a better feature extractor, so that the proposed method can outperform BiC.

Overall, these results indicate that the proposed method is effective to handle catastrophic forgetting in class incremental learning. Our approach not only achieves better performance than state-of-the-art methods but also has a simpler structure.

\begin{table}[t]
\centering
\caption{Class incremental learning performance (top-1 accuracy \%) on CIFAR100 with 2, 5, 10 and 20 incremental steps. The average results over all the incremental steps except the first step are reported. The best results are in bold.}
\setlength{\tabcolsep}{2.5mm}{
\begin{tabular}{c|cccc}
  \toprule
  \#incremental steps & \multicolumn{1}{c|}{2 } & \multicolumn{1}{c|}{5 } & \multicolumn{1}{c|}{10 } & 20  \\
  \midrule
  \midrule
  LwF.MC \cite{li2017learning,rebuffi2017icarl} & \multicolumn{1}{c|}{52.6 } & \multicolumn{1}{c|}{47.1 } & \multicolumn{1}{c|}{39.7 } & 29.7  \\
  iCaRL \cite{rebuffi2017icarl} & \multicolumn{1}{c|}{62.0 } & \multicolumn{1}{c|}{63.3 } & \multicolumn{1}{c|}{61.6 } & 59.7  \\
  EEIL \cite{castro2018end} & \multicolumn{1}{c|}{60.8 } & \multicolumn{1}{c|}{63.7 } & \multicolumn{1}{c|}{63.6 } & \textbf{63.4} \\
  BiC \cite{wu2019large} & \multicolumn{1}{c|}{64.9 } & \multicolumn{1}{c|}{65.1 } & \multicolumn{1}{c|}{63.5 } & 62.1  \\
  Ours  & \multicolumn{1}{c|}{\textbf{65.1 }} & \multicolumn{1}{c|}{\textbf{66.6 }} & \multicolumn{1}{c|}{\textbf{64.5 }} & 62.6  \\
  \midrule
  Upper Bound & \multicolumn{4}{c}{70.1 } \\
  \bottomrule
  \end{tabular}%
}
\label{tab:cifar}%
\end{table}%

\noindent\textbf{Evaluation on CIFAR-100.}
CIFAR-100 has 100 classes, which are divided into 2, 5, 10 and 20 incremental batches respectively in our experiments. The same set of classes in CIFAR-100 are used for all of the compared methods. In CIFAR-100 experiments, we store 2,000 samples in total as the same as previous work. 

The average results over all the incremental steps except the first step are shown in Table \ref{tab:cifar}. Detailed results of all incremental steps are reported in the supplementary material. On CIFAR-100, these methods achieve similar results, which is mainly because this dataset is simple \cite{wu2019large}. Consistent with the results on ImageNet, the proposed method achieves better results compared to state-of-the-art approaches on CIFAR-100 under different settings.

\subsection{Ablation Study}
In this subsection, we analyze the impact of the components of our method. 

\begin{figure}[t]
\centering
\subfloat[impact of restriction to weights]{\includegraphics[width=0.22\textwidth]{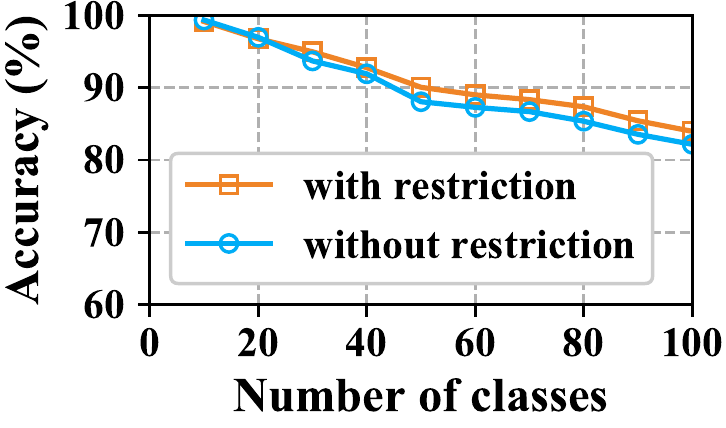}}
\subfloat[impact of norm selection]{\includegraphics[width=0.22\textwidth]{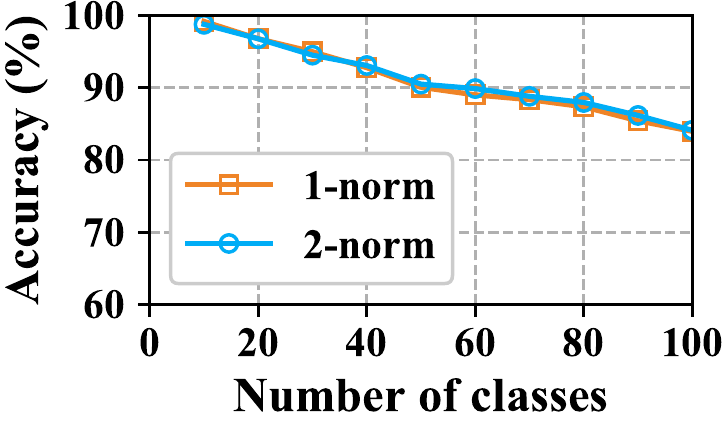}}\\ \vspace{-2mm}
\subfloat[impact of the bias term]{\includegraphics[width=0.22\textwidth]{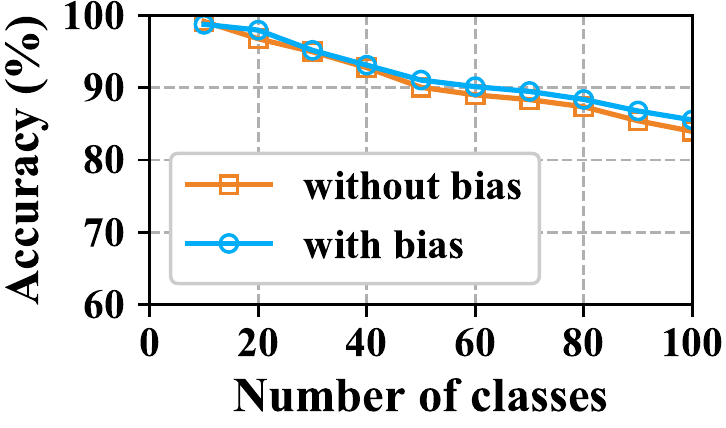}}
\subfloat[impact of exemplar selection]{\includegraphics[width=0.22\textwidth]{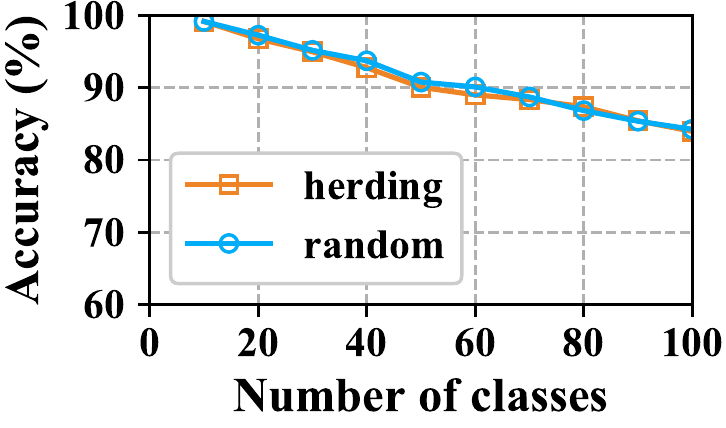}}
\caption{Class incremental learning performance (top-5 accuracy \%) on ImageNet-100 for ablation study.}
\label{fig:ablation}
\end{figure}

\noindent\textbf{Impact of Restriction to the Weights.} We studied the impact of restricting the weights in the FC layer to be positive on ImageNet-100 with 10 incremental steps. As shown in Figure \ref{fig:ablation} (a), our method obtained better performance with restriction to the weights. As discussed in subsection \ref{sec:r_to_w}, this is mainly due to the norms of the weight vectors become more consistent with their corresponding output logits when restricting the weights to be positive, so that the scale factor $\gamma$ obtained by Eq.(\ref{eq:gamma}) is more accurate to suppress the output logits of new classes.

\noindent\textbf{Impact of Norm Selection.} We investigated the impact of different norm used in the proposed method. We compare two norms: 1-norm and 2-norm. Figure \ref{fig:ablation} (b) shows the results. 1-norm and 2-norm achieve similar results, which indicates our method is not sensitive to norm selection.

\noindent\textbf{Impact of the Bias Term in the FC Layer.} We studied the impact of the bias term. With the bias term, the proposed method still calculates the scale factor $\gamma$ by Eq.(\ref{eq:gamma}) based on the weight information and applies it to the output logits for new classes. In other words, the scalar factor $\gamma$ obtained from weight information is used in both the weight term and the bias term in the FC layer. We compare our method with or without using the bias term in the FC layer. Figure \ref{fig:ablation} (c) shows the results. We see that the bias term in the FC layer can only influence the performance slightly.

\noindent\textbf{Impact of Exemplar Selection Strategies.} We investigated the impact of exemplar selection strategies. Random selection and herding selection are considered. Figure \ref{fig:ablation} (d) shows the results. We see that the exemplar selection strategies can only influence the performance slightly.

\section{Conclusions}\label{sec:conclusion}
The goal of class incremental learning is to obtain desirable results on new data, at the same time, retain the previous learned experiences. In this paper, we investigated catastrophic forgetting in class incremental learning. We demonstrated the actual role of knowledge distillation in this problem and the heavily biased weights in the FC layer. We proposed a simple and effective solution to address catastrophic forgetting that maintains the discrimination via knowledge distillation and maintains the fairness via a method called weight aligning. The experimental results on ImageNet-1000, ImageNet-100, and CIFAR-100 show that the proposed method achieves better performance than the previous methods. This work may suggest that there are many useful information hidden in the trained model that is worth exploring.

{\small
\bibliographystyle{ieee_fullname}
\bibliography{mybibliography}
}

\end{document}